\documentclass{svproc}
\usepackage{url}

\usepackage{xcolor}
\usepackage{dsfont}
\usepackage{amsfonts,stmaryrd,amsmath,algorithm,algorithmicx,bbm,placeins,graphicx}
\usepackage{algpseudocode}
\usepackage{caption}
\usepackage{subcaption}
\usepackage[colorlinks=true,linkcolor=blue,citecolor=red,pagebackref]{hyperref}
\usepackage[sort,numbers]{natbib}

\graphicspath{{images/}}

\begin{document}
\mainmatter              % start of a contribution
\title{Non-linear aggregation of filters to improve image denoising}
\titlerunning{Non-linear aggregation of filters to improve image denoising}  % abbreviated title (for running head)
\author{Benjamin Guedj\inst{1} \and Juliette Rengot\inst{2}}

%\author{Anonymous authors\inst{1}}

%
\authorrunning{B. Guedj and J. Rengot} % 

%\authorrunning{Anonymous authors} % 
%abbreviated author list (for running head)
\tocauthor{Benjamin Guedj and Juliette Rengot}

%\tocauthor{Anonymous authors}
%
\institute{Inria, France and University College London, United Kingdom\\
\email{benjamin.guedj@inria.fr} \\ \url{https://bguedj.github.io}
\and
Ecole des Ponts ParisTech, France\\
\email{juliette.rengot@eleves.enpc.fr}}

%\institute{Anonymous affiliation}

\maketitle              % typeset the title of the contribution

\begin{abstract}
We introduce a novel aggregation method to efficiently perform image denoising. Preliminary filters are aggregated in a non-linear fashion, using a new metric of pixel proximity based on how the pool of filters reaches a consensus. We provide a theoretical bound to support our aggregation scheme, its numerical performance is illustrated and we show that the aggregate significantly outperforms each of the preliminary filters.

% We would like to encourage you to list your keywords within
% the abstract section using the \keywords{...} command.
\end{abstract}

\section{Introduction}

\paragraph{}Denoising is a fundamental question in image processing. It aims at improving the quality of an image by removing the parasitic information that randomly adds to the details of the scene. This noise may be due to image capture conditions (lack of light, blurring, wrong tuning of field depth, \dots) or to the camera itself (increase of sensor temperature, data transmission error, approximations made during digitization, \dots). Therefore, the challenge consists in removing the noise from the image while preserving its structure. Many methods of denoising already have been introduced in the past decades -- while good performance has been achieved, denoised images still tend to be too smooth (some details are lost) and blurred (edges are less sharp). Seeking to improve the performances of these algorithms is a very active research topic.
\medskip

The present paper introduces a new approach for denoising images, by bringing to the computer vision community ideas developed in the statistical learning literature. The main idea is to combine different classical denoising methods to obtain several predictions of the pixel to denoise. As each classic method has pros and cons and is more or less efficient according to the kind of noise or to the image structure, an asset of our method is that is makes the best out of each method's strong points, pointing out the "wisdom of the crowd". We adapt the strategy proposed by the algorithm ``COBRA - COmBined Regression Alternative'' \citep{cobra1,cobra2} to the specific context of image denoising. This algorithm has been implemented in the python library \texttt{pycobra}, available on \url{https://pypi.org/project/pycobra/}.
\medskip

Aggregation strategies may be rephrased as collaborative filtering, since information is filtered by using a collaboration among multiple viewpoints. Collaborative filters have already been exploited in image denoising. \cite{dabov2007image} used them to create one of the most performing denoising algorithm: the block-matching and 3D collaborative filtering (BM3D). It puts together similar patches (2D fragments of the image) into 3D data arrays (called ``groups"). It then produces a 3D estimate by jointly filtering grouped image blocks. The filtered blocks are placed again in their original positions, providing several estimations for each pixel. The information is aggregated to produce the final denoised image. This method is praised to well preserve fine details. Moreover, \cite{collaborative2017liu} proved that the visual quality of denoised images can be increased by adapting the denoising treatment to the local structures. They proposed an algorithm, based on BM3D, that uses different non-local filtering models in edge or smooth regions. Collaborative filters have also been associated to neural network architectures, by \cite{strub2015collaborative}, to create new denoising solutions.
\medskip

When several denoising algorithms are available, finding the relevant aggregation has been addressed by several works. \cite{nlmeans2009salmon} focused on the analysis of patch-based denoising methods and shed light on their connection with statistical aggregation techniques. \cite{patch2012chatterjee} proposed a patch-based Wiener filter which exploits patch redundancy. Their denoising approach is designed for near-optimal performance and reaches high denoising quality. Furthermore, \cite{salmon2010agregation} showed that usual patch-based denoising methods are less efficient on edge structures.
\medskip

The COBRA algorithm differs from the aforecited techniques, as it combines preliminary filters in a non-linear way. COBRA has been introduced and analysed by \cite{cobra1}.
\medskip

The paper is organised as follows. We present our aggregation method, based on the COBRA algorithm in \autoref{sec:cobra}. We then provide a thorough numerical experiments section (\autoref{sec:simus}) to assess the performance of our method along with an automatic tuning procedure of preliminary filters as a byproduct.
 
\section{The method}
\label{sec:cobra}

%\subsection{Theoretical scheme}

We now present an image denoising version of the COBRA algorithm \citep{cobra1,cobra2}. For each pixel $p$ of the noisy image $x$, we may call on $M$ different estimators $(f_{1}...f_{M})$. We aggregate these estimators by doing a weighted average on the intensities :
\begin{equation}\label{eq:cobra}
f(p) = \frac{ \sum_{q \in x} \omega(p,q) x(q)}{\sum_{q \in x} \omega(p,q)}\ ,
\end{equation}
and we define the weights as
\begin{equation}\label{eq:weights}
\omega(p,q) = \mathds{1}\left(\sum_{k=1}^{M} \mathds{1}( |f_{k}(p)-f_{k}(q)| \leq \epsilon ) \geq M\alpha\right),
\end{equation}
where $\epsilon$ is a confidence parameter and $\alpha\in (0,1)$ a proportion parameter. Note that while $f$ is linear with respect to the intensity $x$, it is non-linear with respect to each of the preliminary estimators $f_1,\dots,f_M$.
\medskip

These weights mean that, to denoise a pixel $p$, we average the intensities of pixels $q$ such as a proportion at least $\alpha$, of the preliminary estimators $f_1,\dots,f_M$ have the same value in $p$ and in $q$, up to a confidence level $\epsilon$.

%\subsection{Algorithm model}

Let us emphasize here that our procedure averages the pixels' intensities based on the weights (which involve this consensus metric). The intensity predicted for each pixel $p$ of the image is $f(p)$ and the COBRA-denoised image is the collection of pixels $\{f(p), p\in x\}$.
\medskip

\begin{figure}[t]
    \centering
    % \begin{subfigure}[t]{\linewidth}
        \includegraphics[width=\linewidth]{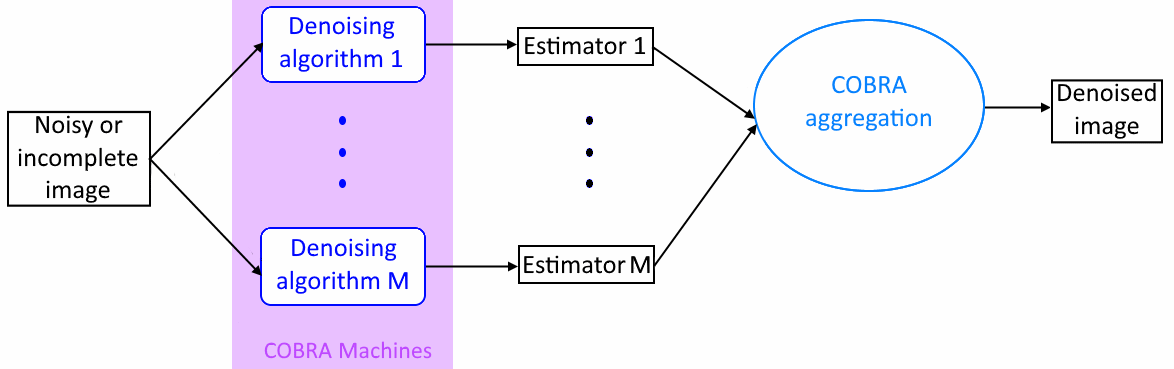}
        \caption{General model}
    % \end{subfigure}
    % \begin{subfigure}[t]{\linewidth}
    %     \includegraphics[width=\linewidth]{images/modele.png}
    %     \caption{Chosen model for numerical experimentation}
    % \end{subfigure}
%  \caption{Scheme of our denoising model}
 \label{fig:modele}
\end{figure}

This aggregation strategy is implemented in the python library \textit{pycobra} \citep{cobra2}. The general scheme is presented in Figure \ref{fig:modele}, and the pseudo-code in Algorithm \ref{fig:pseudoCode}. Users can control the number of used features thanks to the parameter ``$patch\_size$''. For each pixel $p$ to denoise, we consider the image patch, centred on $p$, of size $(2 \cdot patch\_size+1) \times (2 \cdot patch\_size+1)$. In the experiments section, $patch\_size=2$ is usually a satisfying value. Thus, for each pixel, we construct a vector of nine features.

The COBRA aggregation method has been introduced by \cite{cobra1} in a generic statistical learning framework, and is supported by a sharp oracle bound. For the sake of completeness, we reproduce here one of the key theorems.

\begin{theorem}[adapted from Theorem 2.1 in \cite{cobra1}]
\label{theo}
Assume we have $M$ preliminary denoising methods. Let $|x|$ denote the total number of pixels in image $x$. Let $\epsilon \propto |x|^{-\frac{1}{M+2}}$. Let $f^\star$ denote the perfectly denoised image and $\widehat{f}$ denote the COBRA aggregate defined in \eqref{eq:cobra}, then we have
\begin{equation}\label{eq:theo}
\mathbb{E}\left[\widehat{f}(p) - f^\star(p) \right]^2 \leq \underset{m=1,\dots,M}{\min}\, \mathbb{E}\left[ f_m(p)-f^\star(p)\right]^2 + C |x|^{-\frac{2}{M+2}},
\end{equation}
where $C$ is a constant and the expectations are taken with respect to the pixels.
\end{theorem}

What Theorem \ref{theo} tells us is that on average on all the image's pixels, the quadratic error between the COBRA denoised image and the perfectly denoised image is upper bounded by the best (\emph{i.e.}, minimal) same error from the preliminary pool of $M$ denoising methods, up to a term which decays to zero as the number of pixels to the $-1/M$. As highlighted in the numerical experiments reported in the next section, $M$ is of the order of 5-10 machines and this remainder term is therefore expected to be small in most useful cases for COBRA. Note that in \eqref{eq:theo}, the leading constant (in front of the minimum) is 1: the oracle inequality is said to be \emph{sharp}.
Note also that contrary to more classical aggregation or model selection methods, COBRA mactches or outperforms the best preliminary filter's performance, even though it does not need to identify this champion filter. As a matter of fact, COBRA is adaptive to the pool of filters as the champion is not needed in \eqref{eq:cobra}.
More comments on this result, and proofs are presented in \cite{cobra1}.

\begin{algorithm}[t]
    \caption{Image denoising with COBRA aggregation}
    \label{fig:pseudoCode}
    
    \begin{flushleft}
        \textbf{INPUT:} \\
        $im\_noise$ = the noisy image to denoise\\
        $p_\mathrm{size}$ = the pixel patch size to consider\\
        $M$ = the number of COBRA machines to use \\
        \textbf{OUTPUT:} \\
        $Y$ = the denoised image
    \end{flushleft}

    \begin{algorithmic}
        %\Comment{\#\#\# Model Definition \#\#\#}
        \State Xtrain $\gets$ training images with artificial noise
        \State Ytrain $\gets$ original training images (ground truth)
        \State cobra $\gets$ initial COBRA model
        \State cobra $\gets$ to adjust COBRA model parameters with respect to the data (Xtrain, Ytrain)
        \State cobra $\gets$ to load $M$ COBRA machines
        \State cobra $\gets$ to aggregate the predictions
        
        %\Comment{\#\#\# Denoising procedure \#\#\#}
        \State Xtest $\gets$ feature extraction from $im\_noise$ in a vector of size $(nb\_pixels, (2 \cdot p_\mathrm{size} + 1)^2)$
        \State Y $\gets$ prediction of Xtest by cobra
        \State Y $\gets$ to add $im\_noise$ values lost at the borders of the image, because of the patch processing, to Y
    \end{algorithmic}
\end{algorithm}

\section{Numerical experiments}
\label{sec:simus}

This section illustrates the behaviour of COBRA.
All code material (in Python) to replicate the experiments presented in this paper are available at 
\url{https://github.com/bguedj/cobra_denoising}.
%\texttt{<hidden url to preserve anonymity>}.

\subsection{Noise settings}

\begin{figure}[t]
 \centering
 \includegraphics[width=\linewidth]{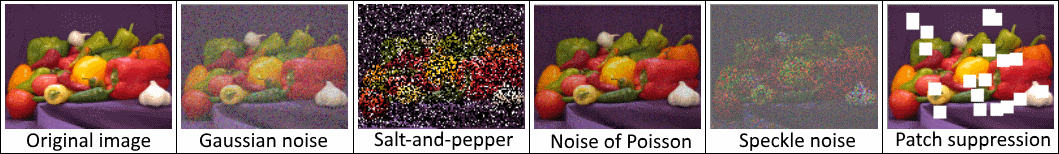}
 \caption{The different kinds of noise used in our experiments.}
 \label{fig:bruit}
\end{figure}

We artificially add some disturbances to good quality images (i.e. without noise). We focus on five classical settings: the Gaussian noise, the salt-and-pepper noise, the Poisson noise, the speckle noise and the random suppression of patches (summarised in Figure \ref{fig:bruit}). 

\subsection{Preliminary denoising algorithms}

We focus on ten classical denoising methods: the Gaussian filter, the median filter, the bilateral filter, Chambolle's method \citep{chambolle2005total}, non-local means \citep{buades2005nonlocal,buades2011nonlocal}, the Richardson-Lucy deconvolution \citep{richardson1972bayesian,lucy1974iterative}, the Lee filter \citep{lee1994speckle}, K-SVD \citep{aharon2006k}, BM3D \citep{dabov2007image} and the inpainting method \citep{damelin2018surface,chuiab2010MRA}. 
%These methods are illustrated in Figure \ref{fig:debruit}. 
This way, we intend to capture different regimes of performance (Gaussian filters are known to yield blurry edges, the median filter is known to be efficient against salt-and-pepper noise, the bilateral filter well preserves the edges, non-local means are praised to better preserve the details of the image, Lee filers are designed to address Synthetic Aperture Radar (SAR) image despeckling problems, K-SVD and BM3D are state-of-the-art approaches, inpainting is designed to reconstruct lost part, etc.), as the COBRA aggregation scheme is designed to blend together machines with various levels of performance and adaptively use the best local method.

\subsection{Model training}

We start with $25$ images $(y_{1}...y_{25})$, assumed not to be noisy, that we use as ``ground truth''. We artificially add noise as described above, yielding {$125$ noisy images $(x_{1}...x_{125})$. 
%Each noisy image is then used to create two independent copies:
Then two independent copies of each noisy image are created by adding a normal noise:
one goes to the data pool to train the preliminary filters, the other one to the data pool to compute the weights defined in \eqref{eq:weights} and perform aggregation. This separation is intended to avoid over-fitting issues \citep[as discussed in][]{cobra1}. The whole dataset creation process is illustrated in Figure \ref{fig:data_set}.

\begin{figure}[t]
    \centering
    \includegraphics[width=0.8\linewidth]{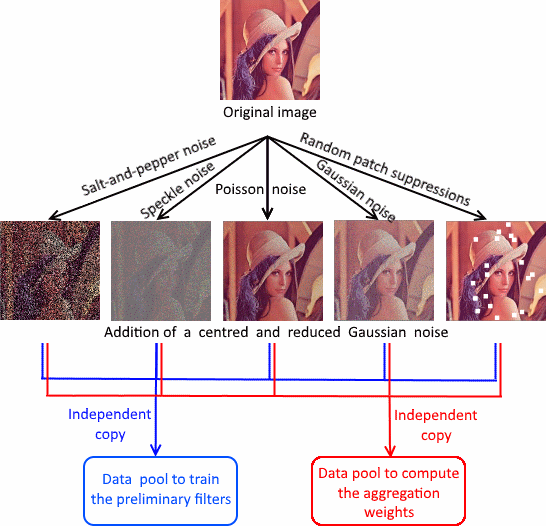}
    %% version couleur : data.png
    %% version NB : dataNB.png
    \caption{Data set construction.}
    \label{fig:data_set}
\end{figure}

\subsection{Parameters optimisation}

The meta-parameters for COBRA are $\alpha$ (how many preliminary filters must agree to retain the pixel) and $\epsilon$ (the confidence level with which we declare two pixels identities similar). For example, choosing $\alpha = 1$ and $\epsilon = 0.1$  means that we impose that all the machines must agree on pixels whose predicted intensities are at most different by a $0.1$ margin.
\medskip

The python library \texttt{pycobra} ships with a dedicated class to derive the optimal values using cross-validation \cite{cobra2}. Optimal values are $\alpha = 4/7$ and $\epsilon = 0.2$ in our setting.

\subsection{Assessing the performance}

We evaluate the quality of the denoised image $I_d$ (whose mean is denoted $\mu_d$ and standard deviation $\sigma_d$) with respect to the original image $I_o$ (whose mean is denoted $\mu_o$ and standard deviation $\sigma_o$) with four different metrics.
\begin{itemize}
 \item Mean Absolute Error (MAE - the closer to zero the better) given by $$\Sigma_{x=1}^{N} \Sigma_{y=1}^{M} \frac{ |I_{d}(x,y)-I_{o}(x,y)| }{N \times M}.$$
 \item Root Mean Square Error (RMSE - the closer to zero the better) given by $$\sqrt{ \Sigma_{x=1}^{N} \Sigma_{y=1}^{M} \frac{ (I_{d}(x,y)-I_{o}(x,y))^{2} }{N \times M}}.$$
 \item Peak Signal to Noise Ratio (PSNR - the larger the better) given by $$10 \cdot \log_{10}\left(\frac{d^{2}}{\mathrm{RMSE}^{2}}\right)$$ with $d$ the signal dynamic (maximal possible value for a pixel intensity).
 \item Universal image Quality Index (UQI - the closer to one the better) given by $$\underbrace{\frac{cov(I_o, I_d)}{\sigma_o \cdot \sigma_d}}_{(i)} \cdot \underbrace{\frac{2 \cdot \mu_o \cdot \mu_d}{\mu_o^2 + \mu_d^2}}_{(ii)} \cdot \underbrace{\frac{2 \cdot \sigma_o \cdot \sigma_d}{\sigma_o^2 + \sigma_d^2}}_{(iii)} $$
 where term $(i)$ is the correlation, $(ii)$ is the mean  luminance similarity, and $(iii)$ is the contrast similarity \citep[][Eq. 2]{wang2002universal}.
\end{itemize}

\subsection{Results}

Our experiments run on the gray-scale ``lena'' reference image (range 0 - 255). In all tables, experiments have been repeated 100 times to compute descriptive statistics. The green line (respectively, red) identifies the best (respectively, worst) performance. The yellow line identifies the best performance among the preliminary denoising algorithms if COBRA achieves the best performance. The first image is noisy, the second is what COBRA outputs, and the third is the difference between the ideal image (with no noise) and the COBRA denoised image.

\paragraph{Results -- Gaussian noise (Figure \ref{fig:images_gauss}).}

We add to the reference image ``lena'' a Gaussian noise of mean $\mu = 127.5$ and of standard deviation $\sigma = 25.5$. Unsurprisingly, the best filter is the Gaussian filter, and the performance of the COBRA aggregate is tailing when the noise level is unknown. When the noise level is known, COBRA outperforms all preliminary filters. Note that the bilateral filter gives better results than non-local means. This is not surprising: \cite{kumar2013image} reaches the same conclusion for high noise levels.

\begin{figure}[h]
 \centering
 \begin{subfigure}[t]{0.3\linewidth}
 \includegraphics[width=\linewidth]{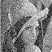}
 \caption{Noisy image}
 \label{fig:gaussNoise}
 \end{subfigure}
 \begin{subfigure}[t]{0.3\linewidth}
 \includegraphics[width=\linewidth]{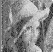}
 \caption{COBRA}
 \label{fig:result_gauss}
 \end{subfigure}
 \begin{subfigure}[t]{0.3\linewidth}
 \includegraphics[width=\linewidth]{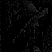}
 \caption{Diff. ideal-COBRA}
 \label{fig:diff_gauss}
 \end{subfigure}
 \begin{subfigure}[t]{0.8\linewidth}
\includegraphics[width=\linewidth]{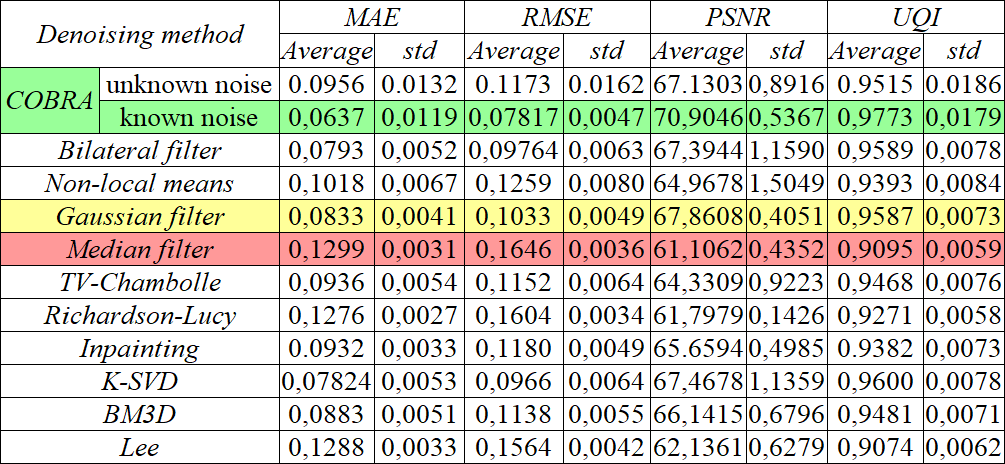}
% \caption{Obtained scores for a Gaussian noise. The experiments have been repeated 100 times to compute statistics.}
 \label{fig:score_gauss}
%  \includegraphics[width=\linewidth]{images/noise_only_gauss.png}
%  \caption{Difference between noisy and denoised images.}
%  \label{fig:noise_only_gauss}
 \end{subfigure}
 \caption{Results -- Gaussian noise.}
 \label{fig:images_gauss}
\end{figure}

\vspace{-1cm}

\begin{figure}[h]
 \centering
 \begin{subfigure}[t]{0.30\linewidth}
 \includegraphics[width=\linewidth]{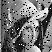}
 \caption{Noisy image}
 \label{fig:spNoise}
 \end{subfigure}
 \begin{subfigure}[t]{0.30\linewidth}
 \includegraphics[width=\linewidth]{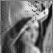}
 \caption{COBRA}
 \label{fig:result_sp}
 \end{subfigure}
 \begin{subfigure}[t]{0.30\linewidth}
 \includegraphics[width=\linewidth]{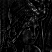}
 \caption{Diff. ideal-COBRA}
 \label{fig:diff_sp}
 \end{subfigure}
 \begin{subfigure}[t]{0.8\linewidth}
 \includegraphics[width=\linewidth]{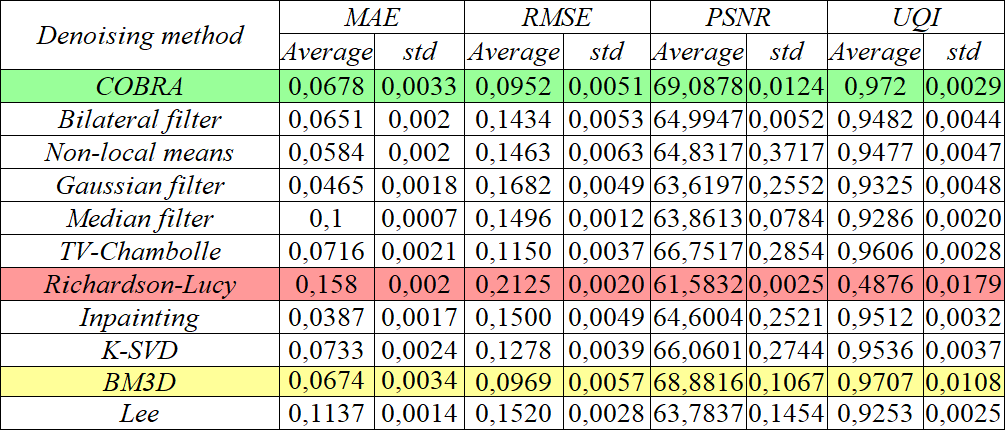}
 %\caption{Obtained scores for a salt-and-pepper noise. The experiments have been repeated 100 times to compute statistics.}
 \label{fig:score_sp}
 \end{subfigure}
 \caption{Result -- salt-and-pepper noise.}
 \label{fig:images_sp}
\end{figure}

% \begin{figure}[ht]
%  \centering
%  \includegraphics[width=.7\linewidth]{images/score_gauss.png}
%  \caption{Obtained scores for a Gaussian noise. The experiments have been repeated 100 times to compute statistics.}
%  \label{fig:score_gauss}
% \end{figure}

% \paragraph{} We have trained the model without using the knowledge we have on the present kind of noise. To improve the results, we repeat the experiment but this time we train the model only on images with Gaussian noise of the same parameters (Figure \ref{fig:result_gauss_connu}). The PSNR increases of  $2.35$ dB (Figure \ref{fig:score_gauss}). Thus, we obtain the best performances.

% \begin{figure}[ht]
%  \centering
%   \begin{subfigure}[t]{0.30\linewidth}
%  \includegraphics[width=\linewidth]{images/result4_gauss_connu.png}
%  \caption{Denoised image with cobra}
%  \label{fig:debruit_gauss_connu}
%  \end{subfigure}
%   \begin{subfigure}[t]{0.3\linewidth}
%  \includegraphics[width=\linewidth]{images/diff_auss_connu.png}
%  \caption{Difference between ideal and denoised images.}
%  \end{subfigure}
%   \begin{subfigure}[t]{0.30\linewidth}
%  \includegraphics[width=\linewidth]{images/noise_only_gauss_connu.png}
%  \caption{Difference between noisy and denoised images.}
%  \label{fig:noise_only_gauss_connu}
%  \end{subfigure}
%  \caption{Obtained results for a known Gaussian noise.}
%  \label{fig:result_gauss_connu}
% \end{figure}

\paragraph{Results -- salt-and-pepper noise (Figure \ref{fig:images_sp}).} The proportion of white to black pixels is set to $sp\_ratio=0.2$ and such that the proportion of pixels to replace is  $sp\_amount=0.1$. 
%As for the Gaussian noise setting, COBRA tails the champion (bilateral filter) when the noise level is unknown and outperforms all filters when it is known (Figure \ref{fig:images_sp}).
Even if the noise level is unknown, COBRA outperforms all filters, even the champion BM3D.

% \begin{figure}[ht]
%  \centering
%  \includegraphics[width=.7\linewidth]{images/score_sp.png}
%  \caption{Obtained scores for a salt-and-pepper noise. The experiments have been repeated 100 times to compute statistics.}
%  \label{fig:score_sp}
% \end{figure}

\paragraph{Results -- Poisson noise (Figure \ref{fig:images_poisson}).}
COBRA outperforms all preliminary filters.

\begin{figure}[t]
 \centering
 \begin{subfigure}[t]{0.30\linewidth}
 \includegraphics[width=\linewidth]{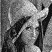}
 \caption{Noisy image}
 \label{fig:poissonNoise}
 \end{subfigure}
 \begin{subfigure}[t]{0.30\linewidth}
 \includegraphics[width=\linewidth]{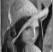}
 \caption{COBRA}
 \label{fig:result_poisson}
 \end{subfigure}
 \begin{subfigure}[t]{0.30\linewidth}
 \includegraphics[width=\linewidth]{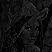}
 \caption{Diff. ideal-COBRA}
 \label{fig:diff_poisson}
 \end{subfigure}
 \begin{subfigure}[t]{0.8\linewidth}
 \includegraphics[width=\linewidth]{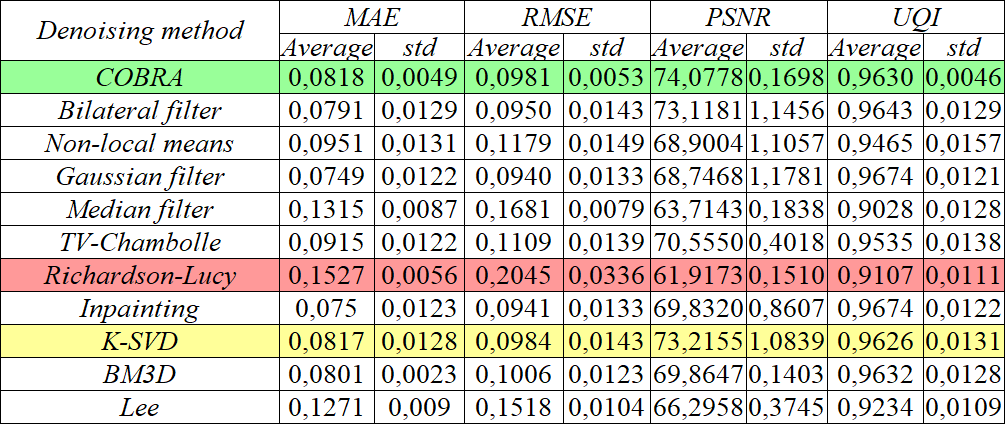}
 %\caption{Obtained scores for a noise of Poisson. The experiments have been repeated 100 times to compute statistics.}
 \label{fig:score_poisson}
 \end{subfigure}
 \caption{Results -- Poisson noise.}
 \label{fig:images_poisson}
\end{figure}

% \begin{figure}[ht]
%  \centering
%  \includegraphics[width=.7\linewidth]{images/score_poisson.png}
%  \caption{Obtained scores for a noise of Poisson. The experiments have been repeated 100 times to compute statistics.}
%  \label{fig:score_poisson}
% \end{figure}

\paragraph{Results -- speckle noise (Figure \ref{fig:images_speckle}).}
When confronted with a speckle noise, COBRA outperforms all preliminary filters. Note that this is a difficult task and most filters have a hard time denoising the image. The message of aggregation is that even in adversarial situations, the aggregate (strictly) improves on the performance of the preliminary pool of methods.

\begin{figure}[t]
 \centering
 \begin{subfigure}[t]{0.30\linewidth}
 \includegraphics[width=\linewidth]{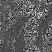}
 \caption{Noisy image}
 \label{fig:speckleNoise}
 \end{subfigure}
 \begin{subfigure}[t]{0.30\linewidth}
 \includegraphics[width=\linewidth]{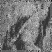}
 \caption{COBRA}
 \label{fig:result_speckle}
 \end{subfigure}
 \begin{subfigure}[t]{0.30\linewidth}
 \includegraphics[width=\linewidth]{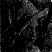}
 \caption{Diff. ideal-COBRA}
 \label{fig:diff_speckle}
 \end{subfigure}
 \begin{subfigure}[t]{0.8\linewidth}
 \includegraphics[width=\linewidth]{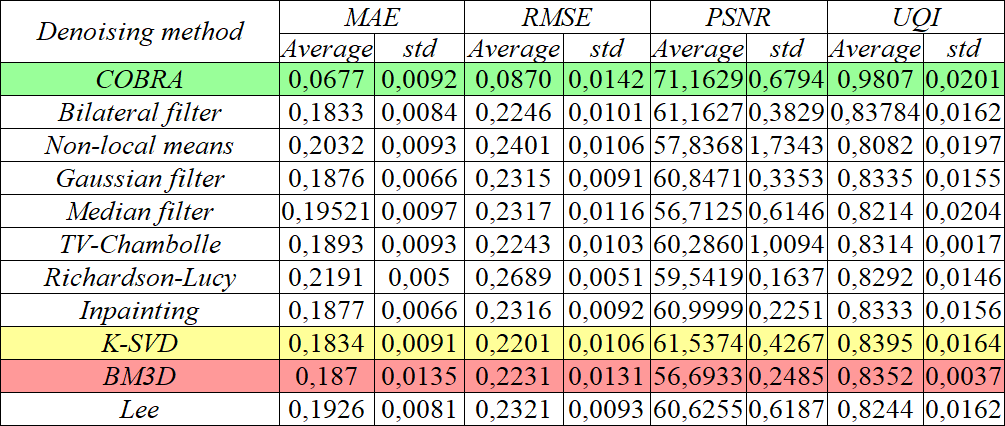}
 %\caption{Obtained scores for a speckle noise. The experiments have been repeated 100 times to compute statistics.}
 \label{fig:score_speckle}
 \end{subfigure}
 \caption{Results -- speckle noise.}
 \label{fig:images_speckle}
\end{figure}

% \begin{figure}[ht]
%  \centering
%  \includegraphics[width=.7\linewidth]{images/score_speckle.png}
%  \caption{Obtained scores for a speckle noise. The experiments have been repeated 100 times to compute statistics.}
%  \label{fig:score_speckle}
% \end{figure}

\paragraph{Results -- random patches suppression (Figure \ref{fig:images_suppr}).}

We randomly suppress 20 patches of size $(4 \times 4)$ pixels from the original image. These pixels become white. Unsurprisingly, the best filter is the inpainting method -- as a matter of fact this is the only filter which succeeds in denoising the image, as it is quite a specific noise.

\begin{figure}[t]
 \centering
 \begin{subfigure}[t]{0.30\linewidth}
 \includegraphics[width=\linewidth]{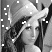}
 \caption{Noisy image}
 \label{fig:supprNoise}
 \end{subfigure}
 \begin{subfigure}[t]{0.30\linewidth}
 \includegraphics[width=\linewidth]{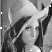}
 \caption{COBRA}
 \label{fig:result_suppr}
 \end{subfigure}
 \begin{subfigure}[t]{0.30\linewidth}
 \includegraphics[width=\linewidth]{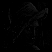}
 \caption{Diff. ideal-COBRA}
 \label{fig:diff_suppr}
 \end{subfigure}
 \begin{subfigure}[t]{0.8\linewidth}
 \includegraphics[width=\linewidth]{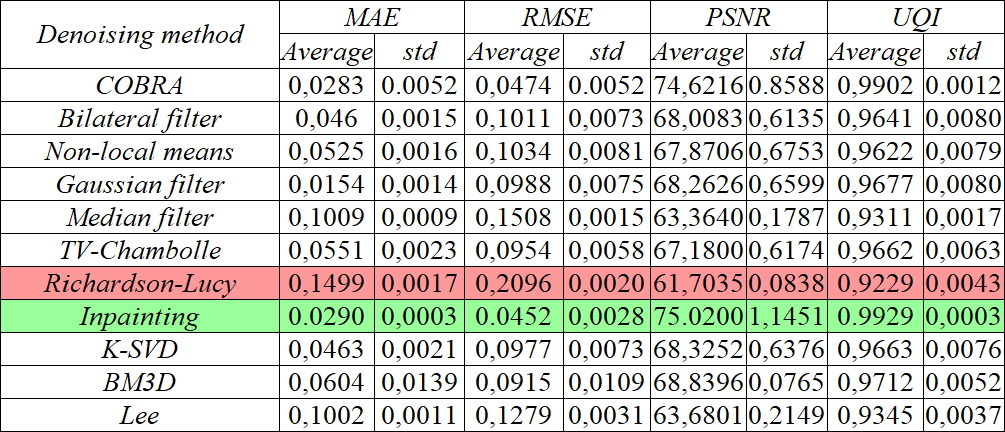}
 %\caption{Obtained scores for a random suppression of patchs. The experiments have been repeated 100 times to compute statistics.}
 \label{fig:score_suppr}
 \end{subfigure}
 \caption{Results -- random suppression of patches.}
 \label{fig:images_suppr}
\end{figure}

% \begin{figure}[ht]
%  \centering
%  \includegraphics[width=.7\linewidth]{images/score_suppr.png}
%  \caption{Obtained scores for a random suppression of patchs. The experiments have been repeated 100 times to compute statistics.}
%  \label{fig:score_suppr}
% \end{figure}

\paragraph{Results -- images containing several kinds of noise (Figure \ref{fig:multi_debruit}).}
On all previous examples, COBRA matches or outperforms the performance of the best filter for each kind of noise (to the notable exception of missing patches, where inpainting methods are superior). Finally, as the type of noise is usually unknown and even hard to infer from images, we are interested in putting all filters and COBRA to test when facing multiple types of noise levels.
We apply a Gaussian noise in the upper left-hand corner, a salt-and-pepper noise in the upper right-hand corner a noise of Poisson in the lower left-hand corner and a speckle noise in the lower right-hand corner. In addition, we randomly suppress small patchs on the whole image (see Figure \ref{fig:multiNoise}).
\medskip

In this now much more adversarial situation, none of the preliminary filters can achieve proper denoising. This is the kind of setting where aggregation is the most interesting, as it will make the best of each filter's abilities. As a matter of fact, COBRA significantly outperforms all preliminary filters.

\begin{figure}[t]
 \centering
 \begin{subfigure}[t]{0.20\linewidth}
     \includegraphics[width=\linewidth]{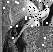}
     \caption{Noisy image}
     \label{fig:multiNoise}
 \end{subfigure}
 
 \begin{subfigure}[t]{0.20\linewidth}
     \includegraphics[width=\linewidth]{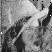}
     \caption{COBRA (Unknown noise)}
 \end{subfigure} 
 \begin{subfigure}[t]{0.20\linewidth}
     \includegraphics[width=\linewidth]{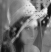}
     \caption{COBRA (Known noise)}
 \end{subfigure} 
 \begin{subfigure}[t]{0.20\linewidth}
     \includegraphics[width=\linewidth]{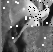}
     \caption{Bilateral filter}
 \end{subfigure}
 \begin{subfigure}[t]{0.20\linewidth}
    \includegraphics[width=\linewidth]{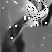}
    \caption{Non-local means}
 \end{subfigure}
 \begin{subfigure}[t]{0.20\linewidth}
     \includegraphics[width=\linewidth]{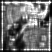}
     \caption{Richardson-Lucy deconvolution}
 \end{subfigure} 
 \begin{subfigure}[t]{0.20\linewidth}
     \includegraphics[width=\linewidth]{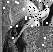}
     \caption{Gaussian filter}
 \end{subfigure}
 \begin{subfigure}[t]{0.20\linewidth}
     \includegraphics[width=\linewidth]{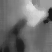}
     \caption{Median filter}
 \end{subfigure}
 \begin{subfigure}[t]{0.20\linewidth}
    \includegraphics[width=\linewidth]{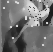}
    \caption{TV Chambolle}
 \end{subfigure}
 \begin{subfigure}[t]{0.20\linewidth}
    \includegraphics[width=\linewidth]{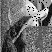}
    \caption{Inpainting}
 \end{subfigure}
 \begin{subfigure}[t]{0.20\linewidth}
    \includegraphics[width=\linewidth]{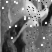}
    \caption{K-SVD}
 \end{subfigure}
 \begin{subfigure}[t]{0.20\linewidth}
    \includegraphics[width=\linewidth]{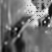}
    \caption{BM3D}
 \end{subfigure}
 \begin{subfigure}[t]{0.20\linewidth}
    \includegraphics[width=\linewidth]{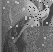}
    \caption{Lee filter}
 \end{subfigure}

 \begin{subfigure}[t]{0.80\linewidth}
 \includegraphics[width=\linewidth]{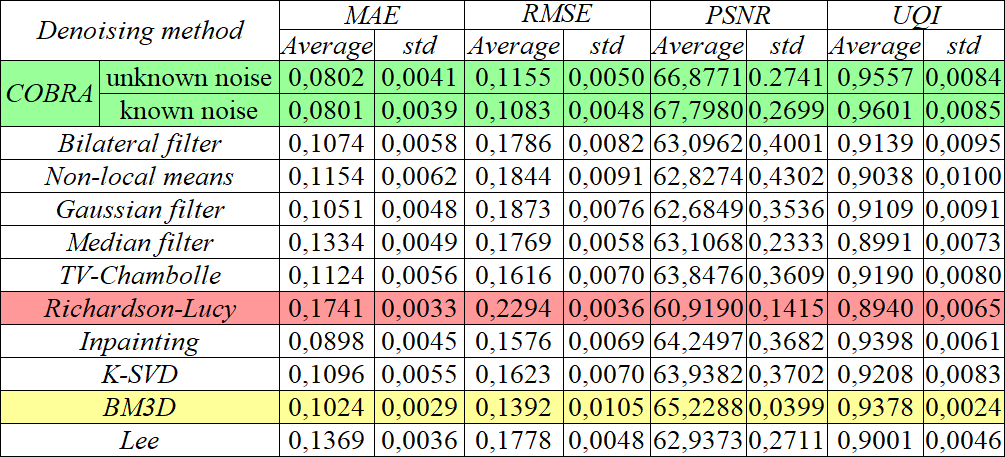}
%  \caption{Obtained scores for a multiple noise. The experiments have been repeated 100 times to compute statistics.}
 \label{fig:score_multi}
 \end{subfigure}
 \caption{Denoising an image afflicted with multiple noises types.}
 \label{fig:multi_debruit}
\end{figure}

% \begin{figure}[ht]
%  \centering
%  \begin{subfigure}[t]{0.40\linewidth}
%  \includegraphics[width=\linewidth]{images/result1_multi.png}
%  \caption{Denoised image with cobra}
%  \label{fig:result_multi}
%  \end{subfigure}
%  \begin{subfigure}[t]{0.40\linewidth}
%  \includegraphics[width=\linewidth]{images/diff_multi.png}
%  \caption{Difference between ideal and denoised images.}
%  \label{fig:diff_multi}
%  \end{subfigure}
%  \caption{Obtained results for a multiple noise.}
%  \label{fig:images_multi}
% \end{figure}

% \begin{figure}[ht]
%  \centering
%  \includegraphics[width=.7\linewidth]{images/score_multi.png}
%  \caption{Obtained scores for a multiple noise. The experiments have been repeated 100 times to compute statistics.}
%  \label{fig:score_multi}
% \end{figure}

% \begin{figure}[ht]
%  \centering
%   \begin{subfigure}[t]{0.40\linewidth}
%  \includegraphics[width=\linewidth]{images/result1_multi_connu.png}
%  \caption{Denoised image with COBRA.}
%  \end{subfigure}
%   \begin{subfigure}[t]{0.40\linewidth}
%  \includegraphics[width=\linewidth]{images/diff_multi_connu.png}
%  \caption{Difference between ideal and denoised images.}
%  \end{subfigure}
%  \caption{Obtained scores for a known multiple noise. The experiments have been repeated 100 times to compute statistics.}
%  \label{fig:result_multi_connu}
% \end{figure}

\subsection{Automatic tuning of filters}

Clearly, internal parameters for the classical preliminary filters may have a crucial impact. For example, the median filter is particularly well suited for salt-and-pepper noise, although the filter size has to be chosen carefully as it should grow with the noise level (which is unknown in practice). A nice byproduct of our aggregated scheme is that we can also perform automatic and adaptive tuning of those parameters, by feeding COBRA with as many machines as possible values for these parameters. Let us illustrate this on a simple example: we train our model with only one classical method but with several values of the parameter to tune. For example, we can define three machines applying median filters with different filter sizes : 3, 5 or 10. Whatever the noise level
%(we test $sp\_amount=0.1$, $sp\_amount=0.3$ and $sp\_amount=0.5$),
our approach achieves the best performance (Figure \ref{fig:result_param_median}). This casts our approach onto the adaptive setting where we can efficiently denoise an image regardless of its (unknown) noise level.

\begin{figure}[t]
 \centering
 \includegraphics[width=\linewidth]{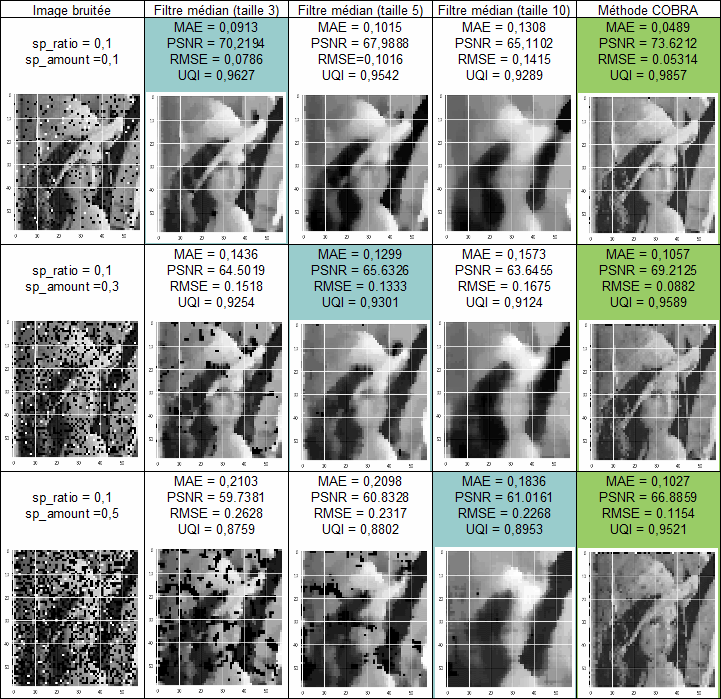}
 \caption{Automatic tuning of the median filter using COBRA.}
 \label{fig:result_param_median}
\end{figure}

\section{Conclusion}

We have presented a generic aggregated denoising method---called COBRA---which improves on the performance of preliminary filters, makes the most of their abilities (e.g., adaptation to a particular kind of noise) and automatically adapts to the unknown noise level. COBRA is supported by a sharp oracle inequality demonstrating its optimality, up to an explicit remainder term which quickly goes to zero.
 Numerical experiment suggests that our method achieves the best performance when dealing with several types of noise. Let us conclude by stressing that our approach is generic in the sense that \emph{any} preliminary filters could be aggregated, regardless of their nature and specific abilities.

\bibliographystyle{bibtex/splncs_srt}
\bibliography{biblio}

\end{document}